\def\arxivversion{1}
\documentclass[letterpaper, 10pt, conference]{ieeeconf}

\IEEEoverridecommandlockouts   
\usepackage{amsmath}
\usepackage{amssymb}

\usepackage{graphicx}
\usepackage{stfloats}
\graphicspath{{figures/}}
\usepackage{booktabs}     
\usepackage{multirow}
\usepackage{microtype}    
\usepackage{url}          
\usepackage{cite}         
\usepackage{tikz}         
\usetikzlibrary{arrows.meta,positioning,fit}

\newif\ifanonymous

\newif\ifdraft
\drafttrue

\usepackage{color}
\definecolor{todored}{rgb}{0.75,0.10,0.10}
\definecolor{noteblue}{rgb}{0.10,0.25,0.70}
\ifdraft
  \newcommand{\todo}[1]{\textcolor{todored}{\textbf{[TODO: #1]}}}
  \newcommand{\note}[1]{\textcolor{noteblue}{\textbf{[note: #1]}}}
\else
  \newcommand{\todo}[1]{}
  \newcommand{\note}[1]{}
\fi

\newcommand{\figref}[1]{Fig.~\ref{#1}}
\newcommand{\tabref}[1]{Table~\ref{#1}}
\newcommand{\secref}[1]{Section~\ref{#1}}

\newcommand{\eqnref}[1]{(\ref{#1})}

\usepackage{xspace}

\newcommand{\methodname}{TapeSim}            
\newcommand{\method}{\textsc{\methodname}\xspace}

\newcommand{\vecx}{\mathbf{x}}

\newcommand{\thick}{\xi}                     
\newcommand{\dhat}{\hat{d}}                  
\newcommand{\adhOpen}{a_n}                  
\newcommand{\adhSlip}{a_t}                  
\newcommand{\adhPress}{p}                   
\newcommand{\adhPressRef}{p_b}              
\newcommand{\adhWork}{W}                    
\newcommand{\adhRelax}{\eta}                
\newcommand{\adhGrowth}{\gamma}             
\newcommand{\adhPressSat}{p_0}              

\newcommand{\freeSet}{\mathcal{F}}           
\newcommand{\clusterSet}{\mathcal{C}}        

\ifdefined\arxivversion
  \usepackage[hidelinks]{hyperref}
\fi

\ifdefined\arxivversion
  \anonymousfalse
\else
  \ifdefined\cameraready
    \anonymousfalse
  \else
    \anonymoustrue
  \fi
\fi

\title{\LARGE \bf \method: Efficient Simulation of Adhesive Tape Dispensing\\ for Robotic Manipulation}

\ifanonymous
  \author{Anonymous Author(s)}
\else
\newcommand{\paperauthors}{Zhaofeng Luo, Xinyu Lu, Jaehoon Choi,
  Zhehuan Chen, Trinity Chung, Xiaowen Qiu, Hugh Nicholas Perkins,
  Gianna Calderon, Alexis Duburcq, Sanghyun Son, Tsun-Hsuan Wang,
  Yi-Ling Qiao, Minchen Li}

\author{%
  \authorblockN{%
    Zhaofeng Luo$^{1,2}$, Xinyu Lu$^{2}$, Jaehoon Choi$^{2}$, Zhehuan Chen$^{2}$\\
    Trinity Chung$^{2}$, Xiaowen Qiu$^{2}$, Hugh Nicholas Perkins$^{2}$, Gianna Calderon$^{2}$\\
    Alexis Duburcq$^{2}$, Sanghyun Son$^{2}$, Tsun-Hsuan Wang$^{2}$, Yi-Ling Qiao$^{2}$, Minchen Li$^{1,2}$}%
  \authorblockA{$^{1}$Carnegie Mellon University \qquad $^{2}$Genesis AI}%
  \ifdefined\arxivversion
    \thanks{This work has been submitted to the IEEE for possible publication.
      Copyright may be transferred without notice, after which this version
      may no longer be accessible.}%
  \fi
}

\fi

\ifdefined\arxivversion
  \hypersetup{%
    pdftitle={\methodname: Efficient Simulation of Adhesive Tape Dispensing for Robotic Manipulation},
    pdfauthor={\paperauthors}%
  }
\fi

\begin{document}
\maketitle
\thispagestyle{empty}
\pagestyle{empty}   

\begin{abstract}
Applying adhesive tape to secure wire harnesses or seal packages requires robots to coordinate a flexible strip, a moving roll, and surfaces that attach and detach. Simulation could make these interactions repeatable for robot development and evaluation, but resolving every adhesive layer is expensive and can suppress roll motion at practical solver tolerances, while a permanently rigid roll cannot release material. We present \method, a tape simulator that concentrates deformation near the unwinding region and along the released strip.
We will release the source code.
A rigid cluster represents most wound material, while an advancing deformable collar enables payout and leaves released tape flexible and reattachable. Optional releasable bonds simplify adhesive interfaces and reduce mean step times for smaller rolls. Controlled swing tests show improved roll rotation.
At 32 turns, clustering gives $3.2$--$3.4\times$ mean
physics-step speedups at a fixed Newton tolerance and
$4.5$--$8.4\times$ for comparable roll motion. Across five real-motion Stick replays, the clustered variants reduce mean image-plane core-landmark error by 23--29\% relative to the full-shell cohesive baseline. On 100 paired Peel cases, they improve balanced accuracy
from 50\% to 72.9--76.3\%, with interface rankings varying across tasks.
A teleoperated box-sealing sequence demonstrates attachment, dispensing, cutting, and sealing in a continuous workflow.
\end{abstract}


\section{Introduction}
\label{sec:intro}

Robotic tape manipulation for wire-harnessing and packaging
requires attaching the free end, dispensing material from a
moving roll, and, where needed, cutting without disturbing
existing attachments. Existing robotic tape-handling
systems~\cite{tushar2025tape,liang2026tape} motivate simulation
tools for developing and evaluating these behaviors.

Simulation supports repeatable policy evaluation for rigid
and soft-object manipulation~\cite{li2025simpler,zhang2026real2sim}.
For tape, it must capture the coupled strip--roll response:
pulling the free end should move and rotate the roll,
release material, and preserve attachment elsewhere.
Errors in these responses can alter a vision-based policy's
observations and task outcome.

Existing models meet only part of these requirements.
Numerical tape placement can exploit a taut strip and
prescribed attachment~\cite{liang2026tape}, while adhesive
contact simulation supports deformation and
peeling~\cite{fang2024adhesion}. A freely moving roll must
also translate, rotate, and unwind. Resolving every wound
layer retains stiff interfaces even when neighboring layers
move together, increasing cost and suppressing motion at
practical solver tolerances. Tighter solves recover motion
at additional cost. Permanently rigidifying the roll removes
this burden but prevents the deformation needed for payout.
We therefore ask where deformation must be retained and
which adhesive interfaces can be simplified.

\begin{figure*}[t]
  \centering
  \includegraphics[width=0.88\textwidth]{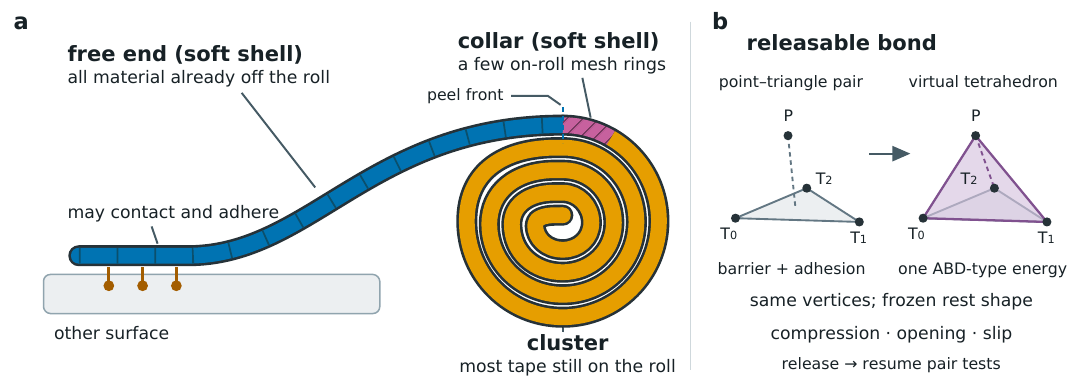}
  \caption{\textbf{Spatial reduction and unified bond interfaces.}
(a) Most on-roll tape forms a rigid cluster, connected through a deformable on-roll collar to the fully released strip, which can adhere to other surfaces. (b) A virtual tetrahedron uses the four vertices of a point--triangle pair, adding no material or degrees of freedom. A single affine-body-type energy replaces its normal-contact barrier, friction, and adhesion; upon release, ordinary pair tests resume. Other pairs remain active. The schematic exaggerates the bond gap.}
  \label{fig:hierarchy}
\end{figure*}

Our key observation is that most wound material moves approximately as a rigid body, while deformation is essential near the advancing release front. We therefore propose \method, which represents most wound tape as a \emph{rigid cluster} connected to the released strip through an advancing deformable \emph{collar}. This allocation retains deformation where needed for progressive payout, while all released tape remains flexible and can adhere to new surfaces (\figref{fig:hierarchy}). Adhesive interfaces can use either the cohesive model or optional releasable \emph{bonds}, which replace selected contact--adhesion pairs with a single interface energy. 
We evaluate roll motion and computational efficiency,
then assess agreement with real robotic interaction through
motion replay and paired real/sim policy rollouts. Our contributions are:
\begin{itemize}
\item A release-front-driven reduction that couples a rigid
wound region to an advancing deformable collar, restoring
shell motion and adhesive history as tape is dispensed.
  \item An optional releasable bond interface, evaluated against cohesive contact to characterize its computational benefits and task-dependent prediction accuracy.
\item A robot-integrated simulator supporting attachment,
dispensing, and cutting, evaluated through real-motion replay,
frozen-policy outcomes, and continuous teleoperated packaging.
\end{itemize}

\section{Related Work}
\label{sec:related}

\textbf{Robotic manipulation of deformable and adhesive objects.}
Robotic manipulation benchmarks cover cables, fabrics, and
bags~\cite{seita2021deformable}, while adhesive tape additionally requires
attachment, release, and continuous dispensing. Tushar \emph{et al.} automate
tape manipulation using specialized tooling~\cite{tushar2025tape}, and Liang
\emph{et al.} learn tape detachment and numerically plan
placement~\cite{liang2026tape} using a taut strip with prescribed attachment.
In contrast, \method simulates a freely moving tape roll coupled to a
progressively released and reattachable strip.

\textbf{Adhesion and contact.}
Tape peeling depends on backing elasticity, peel angle, and fracture
energy~\cite{kendall1975peeling}. Incremental potential contact (IPC) provides
barrier-based contact for deformable solids~\cite{li2020ipc,li2021codim},
with GPU extensions improving
efficiency~\cite{huang2025stiffgipc}. Fang \emph{et al.} extend IPC to adhesive
contact and demonstrate tape peeling~\cite{fang2024adhesion}, while
Raous--Cangemi--Cocou models couple adhesion, friction, and unilateral
contact~\cite{raous1999consistent}. Our focus is complementary: rather than
introducing a more detailed adhesion law, we simplify the dense, persistent
interfaces inside a wound roll using releasable bonds.

\textbf{Reduced simulation for manipulation.}
Affine body dynamics (ABD) compactly represents nearly rigid
objects~\cite{lan2022abd}. Adaptive rigidification switches regions between
rigid and deformable representations for solids and
shells~\cite{mercier2022rigidification,mercier2023shells}, while Embedded IPC
combines reduced deformation with fine collision
geometry~\cite{du2025embedded}. In \method, reduction is tied specifically to
the advancing payout front: most wound material remains rigid, while a
deformable collar advances with release and all dispensed tape remains
deformable. Cutting follows standard mesh-surgery techniques~\cite{pfaff2014tearing}.

\textbf{Real-to-sim robot evaluation.}
SIMPLER aligns simulated control and visual observations for evaluating frozen
robot policies~\cite{li2025simpler}, and Zhang \emph{et al.} extend real-to-sim
evaluation to soft-body interactions~\cite{zhang2026real2sim}. RealSimLoop
uses differentiable reduced-order simulation and visual feedback for online
material adaptation~\cite{cen2026realsimloop}. We instead keep the tape models
fixed and evaluate their predictive fidelity through real-motion replay and
paired real/sim outcomes of frozen policies.
\section{Method}
\label{sec:method}

\subsection{Thin-Shell Simulation with IPC}
\label{sec:problem}
\label{sec:method:overview}

We model the tape backing with a triangular thin shell using StVK membrane elasticity and discrete-shell bending. Both use the flat rest mesh, so winding stores elastic strain rather than defining a new stress-free configuration. We first describe the nonadhesive strip, then add adhesion in \secref{sec:method:adhesion}.

At each time step with size $\Delta t$, we solve jointly for the tape and body
coordinates $\mathbf q$. Without adhesion, the next configuration minimizes
\begin{equation}
 \begin{split}
 \Phi_0(\mathbf q)={}&\mathcal I(\mathbf q)+\mathcal E_{\rm el}(\mathbf q)\\
       &+\mathcal E_{\rm nc}(\mathbf q)+\mathcal E_{\rm fric}(\mathbf q).
 \end{split}
 \label{eq:objective}
\end{equation}
Here $\mathcal I$ penalizes departure from inertially predicted motion,
including external forces. $\mathcal E_{\rm el}$ resists deformation,
$\mathcal E_{\rm nc}$ handles normal contact, and $\mathcal E_{\rm fric}$ resists
sliding along contacting surfaces. Potential-energy terms include scaling of $\Delta t^2$.

Incremental potential contact (IPC) detects nearby vertex--triangle and
edge--edge pairs, both within the strip and against other
objects~\cite{li2020ipc,li2021codim}. Contact half-thickness $\thick$ gives
the mesh a finite collision thickness: equal-thickness midsurfaces touch
at separation $2\thick$. The \emph{barrier} energy $\mathcal E_{\rm nc}$ activates as surfaces
approach and rises without bound as clearance vanishes. Continuous collision detection limits solver updates to prevent surfaces from crossing.
Projected Newton iterations minimize this objective \eqref{eq:objective}; preconditioned
conjugate gradients (PCG) solve the linearized systems \cite{li2026physics}.

\subsection{From Cohesive Contact to Releasable Bonds}
\label{sec:method:adhesion}

\textbf{Cohesive baseline.}
Contact repels approaching surfaces; adhesion must also resist pulling
them apart. Our RCC-inspired baseline assigns each adhesive point--triangle
pair a persistent intensity $\beta\in[0,1]$, from disengaged to fully
engaged. Either tape face can contact objects, but only the sticky face
can adhere. The cohesive model adds $\sum_{a\in\mathcal A}\mathcal E_a$
to \eqnref{eq:objective}. For pair distance $d$, rest gap $d^*$, and
tangential slip $\Delta\vecx_t$ measured in a lagged contact frame,
\begin{equation}
 \mathcal E_a=\Delta t^2\beta^2
 \left[\frac{C_n}{2\dhat}(d-d^*)^2+
       \frac{C_t}{\dhat}\|\Delta\vecx_t\|^2\right].
 \label{eq:adhesion}
\end{equation}
$\dhat$ is the contact activation length in IPC. $C_n,C_t$ are per-pair coefficients in newtons that control opening and sliding resistance. A fixed side mask excludes backing-to-backing adhesion, not contact.
Compression promotes engagement, while opening and slip promote debonding. After each motion solve, active sticky pairs within the contact band update $\beta^{k+1}=\operatorname{clip}_{[0,1]}(\beta^k+\Delta t\,\dot\beta)$.
With $\adhOpen=4[(d-d^*)/\dhat]^2$ and
$\adhSlip=4(C_t/C_n)\|\Delta\vecx_t/\dhat\|^2$, the implemented rate is
\begin{equation}
 \dot\beta=\begin{cases}
 \frac{10}{\adhRelax}[\adhWork-\beta(\adhOpen+\adhSlip)]_-, & \adhPress<0,\\[2pt]
 \frac{100\adhGrowth}{\adhPressRef}[\adhPress-\beta\adhPressSat]_+
 +\frac{10}{\adhRelax}[\adhWork-\beta\adhSlip]_-, & \adhPress\geq0.
 \end{cases}
 \label{eq:beta-update}
\end{equation}
Here $[z]_+=\max(z,0)$ and $[z]_-=\min(z,0)$.
The compression measure is
$\adhPress=-2d\dhat B'(d^2)-(C_n/\dhat)\beta^2(d-d^*)$,
where the IPC kernel $B$ omits time-step and area factors.
$\adhPressRef$ is the barrier-only compression at the contact-band midpoint
above the pair's contact thickness (growth is zero if $\adhPressRef=0$).
$\adhWork,\adhRelax,\adhGrowth,\adhPressSat$ control retention, debonding,
engagement, and pressure saturation, respectively.

Many wound-layer interfaces barely move, yet their stiff barrier
and cohesive terms increase solve cost and can suppress roll
rotation under practical solver stopping tolerances.
Interface histories and cluster membership stay fixed during each motion
solve and update at time-step boundaries.

\textbf{Unified bond response.}
To accelerate cohesive contact, we introduce a persistent bond for selected adhesive point--triangle pairs. The contacting point $\mathbf x_P$ and the triangle's three vertices define a \emph{virtual tetrahedron} (\figref{fig:hierarchy}b), adding no material or degrees of freedom. Its creation-time edge matrix $D_b$ records the reference shape, while $D_s$ describes its current shape. With deformation gradient $F_b=D_sD_b^{-1}$ and reference volume $V_b=|\det D_b|/6$, we use the rigidity energy~\cite{lan2022abd}
\begin{equation}
  \mathcal E_b=\Delta t^2 V_b\kappa_b
              \|F_bF_b^\mathsf T-I\|_F^2.
  \label{eq:bond}
\end{equation}
Stiffness $\kappa_b$ (Pa) resists opening, compression, and slip while allowing joint translation and rotation. The bond replaces the source pair's cohesive energy, normal-contact barrier, and friction, for better computing efficiency; other contact pairs remain active. This defines an alternative finite-stiffness interface law.

  \textbf{Bond creation and release.}
  We update bonds at the end of each time step, keeping the bond set fixed throughout each motion solve. A bond is created when an eligible adhesive point--triangle pair lies within a prescribed bonding distance. Once established, the bond persists until its release criterion is met.   Our primary release criterion is force-based and active only in tension. Let $d$ be the current point--triangle distance and $d_{\rm ref}$ the bond's reference separation. We evaluate the restoring force $\mathbf f_b=-\partial(\mathcal E_b/\Delta t^2)/\partial\mathbf x_P$ and release the bond only when $d>d_{\rm ref}$ and $\|\mathbf f_b\|>f_{\rm break}$. When $d\leq d_{\rm ref}$, this force-based criterion is disabled,
  so compressive loads do not trigger release.
  Here $f_{\rm break}$ is the prescribed force threshold in newtons.
  Removing the bond restores ordinary IPC processing for the source pair
  at the next contact search.

\subsection{Rigid Wound Region and Advancing Collar}
\label{sec:method:hierarchy}

Even with bonds, resolving the deformation of every wound layer remains costly. Away from the release front, neighboring layers largely move together, so we represent most wound tape as a rigid cluster \(\mathcal C\). Near the front, a deformable collar \(\mathcal L\) allows tape to bend away from the roll and release its adhesive support. The released region \(\mathcal F\) also remains a shell, preserving the deformation needed for subsequent contact and attachment to new surfaces (\figref{fig:hierarchy}). The hub remains a separate body coupled to the tape through the tape–hub interface.

Instead of three unknowns per member vertex, the cluster shares six
translation/rotation degrees of freedom. A stored local coordinate $y_i$
maps to world position $x_i=R_p y_i+t_p$ under cluster pose $(R_p,t_p)$.
Member elasticity and internal contacts/adhesion are omitted; boundary
elements and external contacts still couple the cluster to deformable tape
and the scene. To transfer their contributions to reduced coordinates, let
$J=\partial\vecx/\partial\mathbf q$ map coordinate changes to vertex motion.
For vertex-space energy gradient $g_x$ and Hessian $H_x$,
\begin{equation}
 g_q=J^\mathsf T g_x,\qquad
 H_q=J^\mathsf T H_xJ+
 \sum_j(g_x)_j\nabla_q^2[\vecx]_j.
 \label{eq:relocation}
\end{equation}
The sum runs over scalar vertex coordinates and accounts for the nonlinear rotation map.
Collision detection uses conservative advancement along screw trajectories,
bounding their deviation from straight chords during line search.

Mass contributions from member elements are assigned to the cluster, while shell vertices retain contributions from incident elements outside the cluster. The elastic rest geometry is preserved during this change of representation. At cluster formation, vertex velocities are projected onto rigid motion. When a cluster vertex becomes deformable, it inherits \(v_i=v_p+\omega_p\times(x_i-c_p)\), where \(c_p\) is the cluster center of mass and \(v_p,\omega_p\) are its instantaneous linear and angular velocities.

\textbf{Front update.}
A fixed cluster boundary would prevent continued payout, so we maintain a collar of \(N\) mesh-graph edges between the cluster and free region. Strip ordering and adhesive support initialize cluster membership, excluding the collar and free end. Every tape vertex stores its initial local position \(y_i\) in the cluster frame. Its carried position \(\tilde x_i=R_p y_i+t_p\) follows rigid cluster motion, while collar positions \(x_i\) remain independent shell unknowns.

After each motion solve, we update interface histories and test collar vertices one graph edge from the previous free region \(\mathcal F_k\). The gate \(g_i=\|x_i-\tilde x_i\|>\delta\) detects deformation away from rigid cluster motion. With bond interfaces, candidates must also be marked by a release event and have no remaining live support; with cohesive interfaces, the gate does not require $\beta$ to reach zero. Passing vertices join \(\mathcal F_{k+1}\). We then recompute graph distance to this set and restore deformation to cluster elements with a corner less than \(N\) edges away, replenishing the collar for subsequent payout. Thus the free region grows as the cluster shrinks:

\begin{equation}
 \freeSet_k\subseteq\freeSet_{k+1},\qquad
 \clusterSet_{k+1}\subseteq\clusterSet_k.
 \label{eq:front}
\end{equation}
Candidate adjacency is evaluated against the previous free region $\mathcal F_k$, preventing releases within one update from recursively propagating through the collar. As cluster elements become deformable, cohesive interfaces recover
the intensities $\beta$ saved during winding, while bond interfaces
recover their original bonds and reference shapes. Released tape may adhere elsewhere but never rejoins the original cluster.

\subsection{Free-Strip Cutting}
\label{sec:method:cutting}
\label{sec:method:impl}

Cutting is triggered by the peak local pressure from blade--tape contact:
\begin{equation}
  p_{\max}=\max_e
  \frac{\sum_{c\rightarrow e}F_{n,c}}{A_e}.
\end{equation}
Here $F_{n,c}$ is the IPC barrier normal force from sharp blade contact $c$ assigned to tape element $e$, including stiffness and area weights. The associated reference area $A_e$ is the triangle area or, for edges and vertices, a share of incident triangle areas. We trigger cutting when $p_{\max}$ reaches the prescribed threshold for two consecutive time steps. At a step boundary, we project the blade curve and nearby triangles onto a local plane, remove a finite-width kerf, and retriangulate the retained material. The kerf separates newly exposed edges beyond their combined contact thickness, avoiding coincident surfaces in the next contact step.

New vertices inherit material and kinematic state through parent-triangle interpolation, with mass determined by retained rest areas. We rebuild shell/contact connectivity and remap grasp constraints so both pieces continue interacting with the scene.


\section{Evaluation}
\label{sec:experiments}

We evaluate tape handling, computational cost, and real/sim
agreement through motion replay and paired policy rollouts.
We also test adhesion sensitivity and demonstrate continuous
teleoperated packaging.

\textbf{Models and shared settings.}
We compare Coh-F, Bond-F, Bond-R, and Coh-R. Coh/Bond
selects cohesive or bond tape--tape and tape--hub interfaces;
F/R selects full shells or rigid clusters.
Within each comparison, all four models share tape geometry
and nonadhesive material parameters, including membrane
and bending stiffness. F/R pairs use the same input motion
or frozen policy. Tape--hub parameters for each adhesive
model are shared across Peel, Stick, and Wrap.
All four Peel models use the calibrated cohesive tabletop
interface (\secref{sec:exp:physical}).

Unless noted otherwise, $\Delta t=0.01$~s and the Newton
velocity tolerance is $10^{-2}$~m/s.
Payout, fixed-tolerance timing, Stick replay, and cutting
use $10^{-3}$~m/s; the comparable-motion benchmark selects
tolerances by model and roll size.
Blade indentation uses a 5-ms step and box packaging a
$1/30$-s step. Reduced rolls use an $N=10$-edge collar.
Initial placements are manually verified; real-robot test
outcomes are not used to select simulator configurations.

\subsection{Physical Setup and Tape-Handling Capabilities}
\label{sec:exp:setup}
\label{sec:exp:physical}

\begin{figure*}[t]
  \centering
  \begin{minipage}[t]{.35\textwidth}
    \vspace{0pt}
    \includegraphics[width=\linewidth]{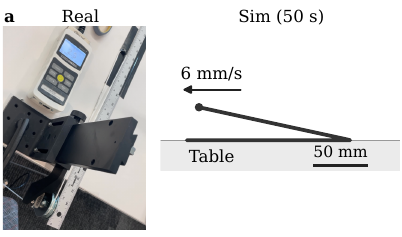}
  \end{minipage}
  \begin{minipage}[t]{.54\textwidth}
    \vspace{0pt}
    \includegraphics[width=\linewidth]{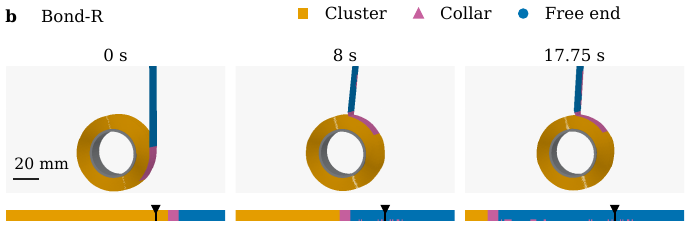}
  \end{minipage}
  \par
  \includegraphics[width=0.9\textwidth]{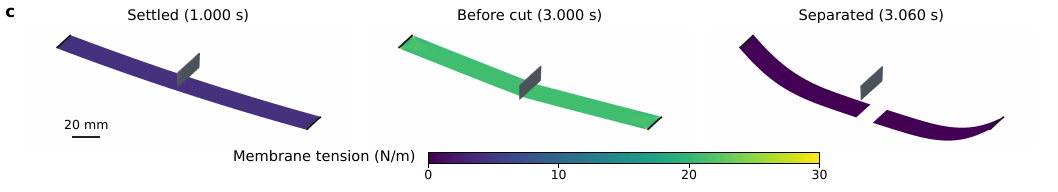}
  \caption{\textbf{Peel-force calibration and tape-handling feasibility.}
(a) Force-gauge measurement and simulated peel shape at 50 s; the simulated fixture gives a median reaction of 2.54 N at \(C_n=C_t=0.675\). The arrow indicates pulling direction; the simulated
profile uses equal horizontal and vertical scales.
(b) A 32-turn Bond-R roll dispenses tape as the collar advances.
Orange/magenta/blue indicate cluster/collar/free end, and gray the hub.
State bars span the same rest-material interval; black markers track one
material row. Roll views share a common scale.
(c) Blade indentation and separation of a fixed-end strip.
Black edges are fixed and gray denotes the blade. Color denotes positive
maximum principal membrane tension (N/m), showing tension release after
contact-pressure-triggered cutting.}
  \label{fig:capabilities}
\end{figure*}

\subsubsection{Tape Assets}
Except for box packaging (\secref{sec:exp:showcase}),
experiments use the 19-mm tape assets in
\tabref{tab:tape-parameters}. Mass is computed from measured
density, flat rest area, and full thickness.
\begin{table}[t]
  \centering
  \caption{Parameter settings for the 19-mm tape assets.}
  \label{tab:tape-parameters}
\begin{tabular}{@{}lll@{}}
  \toprule
  Quantity & Value & Source \\
  \midrule
  Tape width & 19 mm & Measurement \\
  Full film thickness & 0.18 mm & Measurement \\
  Core inner / outer diam. & 38.0 / 42.2 mm & Measurement \\
  Core axial height & 20.0 mm & Measurement \\
  Tape density & 1300 kg/m$^3$ & Measurement \\
  Membrane / bending $E$ & 50 / 50 MPa & Configured \\
  Poisson ratio & 0.45 & Configured \\
  Tape mass per length & 4.446 g/m & Derived \\
  Contact band $\dhat$ & 0.18 mm & Configured \\
Table $C_n$ / $C_t$ (N) & 0.675 / 0.675 & Numerical fit \\
  \bottomrule
\end{tabular}

\end{table}
We wind a triangular shell with six cells across its width,
adhesive face inward, around a hollow core. After settling
in zero gravity, we save its geometry and adhesive interfaces
for full or reduced simulation.
Our robot platform uses Marvin arms equipped with Wuji hands.
For paired real/sim trials, simulated scenes reconstruct
the initial geometry from the first real frame and include
a movable core and the tabletop or target stand.

\subsubsection{Peel-force calibration}
We calibrate tabletop release against a measured peel force of 2.5~N. In both real and simulated fixtures, tape is adhered to the tabletop and its free end is lifted and pulled at approximately $180^\circ$. The real free end is attached to a force gauge. In simulation, a 300-by-19-mm strip is pulled at 6~mm/s, and we measure the pull-direction reaction during advancing release. The simulated robot fixture gives a median of 2.54~N at $C_n=C_t=0.675$. These coefficients are fixed for all four Peel models before policy evaluation.

\subsubsection{Dispensing and cutting feasibility}
The collar follows the release front while maintaining strip continuity
(\figref{fig:capabilities}b). With the tip held in zero gravity, we
translate the freely rotating hub at 25~mm/s. During 17.75~s of payout,
the 32-turn Bond-R boundary advances 370.5~mm along the rest-material
coordinate. The tracked row passes from cluster through collar to free
end: newly released material becomes deformable as dispensing proceeds.

To demonstrate cutting feasibility, a rigid blade descends at 6~mm/s
onto a fixed-end 200-by-19-mm strip (\figref{fig:capabilities}c).
We prescribe a peak local contact-pressure threshold of 1~kPa,
sustained for two consecutive time steps, using the same pressure threshold
and two-step trigger as the teleoperated box-packaging demonstration
(\secref{sec:exp:showcase}). The cut separates the strip and releases
its tension; the matched no-cut control remains connected through 6.5~s.

\subsection{Roll Motion and Computational Efficiency}
\label{sec:exp:dynamics}
\label{sec:exp:pareto}

A held-tip swing isolates the roll's response to motion of the free end.
We lift and hold the tip, drive the roll center through a 30$^\circ$
arc, and release it for 10 s, leaving roll rotation unconstrained.
We test all four models at 2, 8, 16, 24, and 32 turns
over a 17.5-s sequence. The additional comparison for
comparable roll motion uses 2, 8, 16, and 32 turns.
These rolls use $C_n=C_t=1$ N with
$\adhWork=1$, $\adhRelax=100$, $\adhGrowth=1$, and $\adhPressSat=0$
for cohesive interfaces, or $\kappa_b=10^6$ Pa bonds with a 0.5-N release threshold and no rest-gap
snapping. The front uses the Euclidean gate with
$\delta=5\dhat=0.9$ mm. 
For these benchmarks, Newton iterations stop when
$\max_i\|\Delta\mathbf{x}_i\|_\infty/\Delta t<\varepsilon_v$.
Here $\Delta\mathbf{x}_i$ is the world-space Newton update
before CCD and line-search scaling. For R models,
translation and rotation updates are mapped to constituent
vertices, including clustered tape vertices.
With the tip held, gravity loads the strip, whose off-center
pull turns the roll. In this approximately planar test,
we characterize roll motion by the hub angle $\theta(t)$
about the roll axis and its range
$\Delta\theta=\max_t\theta(t)-\min_t\theta(t)$ over
the 10-s release.

\begin{figure*}[!t]
  \centering
  \includegraphics[width=0.9\textwidth]{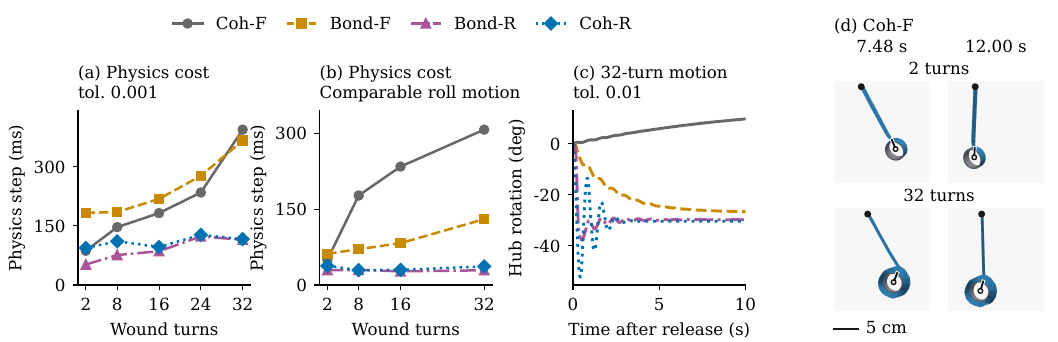}
\caption{\textbf{Roll motion and computational cost.}
(a) Mean physics-step time over 1750 steps at Newton
tolerance $10^{-3}$~m/s, excluding cluster updates
and rendering.
(b) Mean physics-step time for comparable roll motion,
using tolerances selected by matching hub-rotation ranges.
(c) Hub rotation after release for a 32-turn roll at
$10^{-2}$~m/s, relative to the first post-release state.
(d) Coh-F at 7.48 and 12~s, before and after hub release
at 7.5~s. Views share time and scale; dots mark the held tip,
circles the hub center, and spokes the hub orientation.}
  \label{fig:dynamics}
\end{figure*}

\textbf{Rotation improves when internal interfaces are simplified.}
Tightening the 32-turn Coh-F Newton tolerance from $10^{-2}$
to $10^{-3}$~m/s raises its angular range from $9.7^\circ$
to $38.1^\circ$, with the asset, interface law, time step,
and input unchanged. This identifies a solver-induced
component of the suppressed motion. At $10^{-2}$~m/s,
Coh-F's range falls from $28.1^\circ$ at 2 turns to
$9.7^\circ$ at 32 turns. At 32 turns, bonds increase the
range to $26.6^\circ$; clustering yields $37.7^\circ$
for Bond-R and $52.2^\circ$ for Coh-R
(\figref{fig:dynamics} c,d).

\textbf{Clustering provides the main computational savings.}
The fixed-tolerance benchmark uses $10^{-3}$~m/s.
Because a common stopping tolerance can yield different roll
responses, we also compare costs for comparable motion below.
Timing uses the same RTX PRO 6000 Blackwell Server Edition
GPU within each group, with two independent 32-turn repeats.
Physics steps are bracketed by GPU synchronization, excluding
front updates, control, telemetry, and rendering.
At 32 turns, clustering reduces mean physics-step time from
366 to 115~ms within the bond family, a $3.2\times$ speedup;
the cohesive family gains $3.4\times$
(\figref{fig:dynamics} a).
Including front updates retains a $3.1\times$ bond-family speedup.

At this tolerance, Bond-R is approximately $1.8\times$
and $1.5\times$ faster than Coh-R for 2- and 8-turn rolls,
respectively. Their mean step times are nearly equal at
32 turns (115 versus 116~ms).

\textbf{Cost for comparable roll motion.}
For each roll size, Bond-R at $10^{-2}$~m/s defines
$\Delta\theta^\star$. We select Newton tolerances satisfying
$|\Delta\theta-\Delta\theta^\star|/\Delta\theta^\star\leq0.05$.
The selected runs exhibit comparable roll trajectories
and velocities; \figref{fig:dynamics}b reports their
mean physics-step times.
At 32 turns, clustering reduces mean step time from 307.77
to 36.73~ms in the cohesive family and from 130.43 to
29.14~ms in the bond family, giving $8.4\times$ and
$4.5\times$ speedups, respectively.
Under this criterion, Bond-R is $1.26\times$ faster
than Coh-R at 32 turns.

In a separate 32-turn comparison, Bond-F and Bond-R use
$10^{-3}$~m/s, with Bond-F at $10^{-4}$~m/s as the
numerical reference and the same time step.
Clustering reduces roll-center RMSE from 11.11 to 5.09~mm
and orientation RMSE from $4.68^\circ$ to $2.17^\circ$.

\subsection{Open-Loop Motion Fidelity}
\label{sec:exp:replay}

\begin{figure*}[t]
  \centering
  \includegraphics[width=0.87\textwidth]{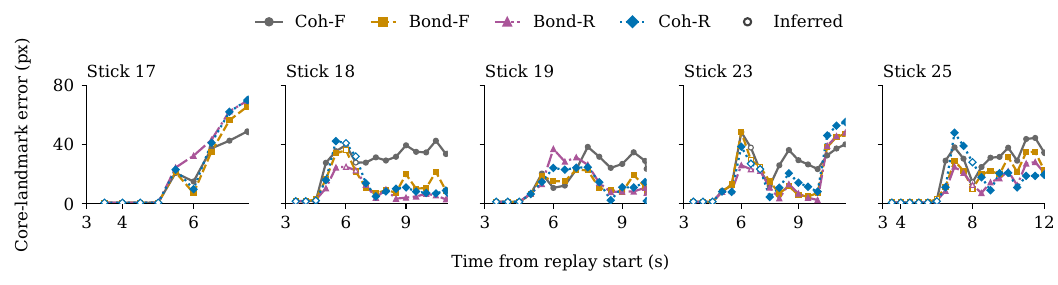}
  \includegraphics[width=0.9\textwidth]{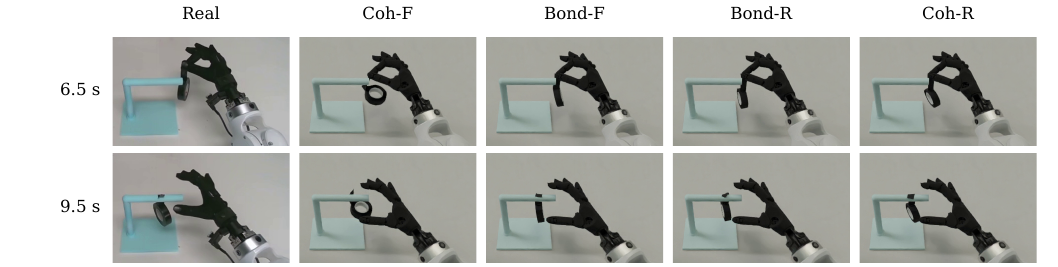}
\caption{\textbf{Clustered models retain Stick replay fidelity.}
Top: ego-camera core-landmark error for five real-motion replays;
each curve averages two simulation repeats. Time is measured from
recording start, excluding the stationary first 3~s. Hollow markers
identify estimated real landmarks.
Bottom: synchronized real/sim frames from Stick 19, repeat 1,
at 6.5 and 9.5~s, using the same crop across models.}
  \label{fig:replay}
\end{figure*}

Stick replay tests tape motion as the robot grasps the free end,
lifts the roll off the tabletop, and attaches the end to a stand.
We replay measured arm and finger trajectories from five real
Stick episodes, twice per model, with shared geometry and cameras.

We measure Euclidean core-landmark error in undistorted
$848\times480$ ego-camera images. Manually verified real
landmarks use the inner-hole ellipse center or, in side
views, the outer-core midpoint excluding the strip.
Partially obscured points are estimated from visible
contours; all models and repeats share these annotations.
Simulated landmarks use the projected camera-facing
inner-rim ellipse center or the projected center of the
hub's three-dimensional axis-aligned bounding box,
respectively. Timestamps select the nearest simulated state
without fitting a time shift. We average errors after
3~s within each run, across repeats, then equally across
the five episodes.

Coh-R and Bond-R reduce mean core-landmark error from Coh-F's
22.96~px to 17.57 and 16.28~px, respectively, corresponding to
reductions of 23\% and 29\% (\figref{fig:replay}). Their similar
mean errors support both interface choices within the clustered
representation. Bond-F gives 17.03~px, indicating that clustering
also retains comparable replay fidelity within the bond family.

\subsection{Closed-Loop Prediction of Policy Outcomes}
\label{sec:exp:policy}

\begin{figure*}[t]
  \centering
  \includegraphics[width=0.85\textwidth]{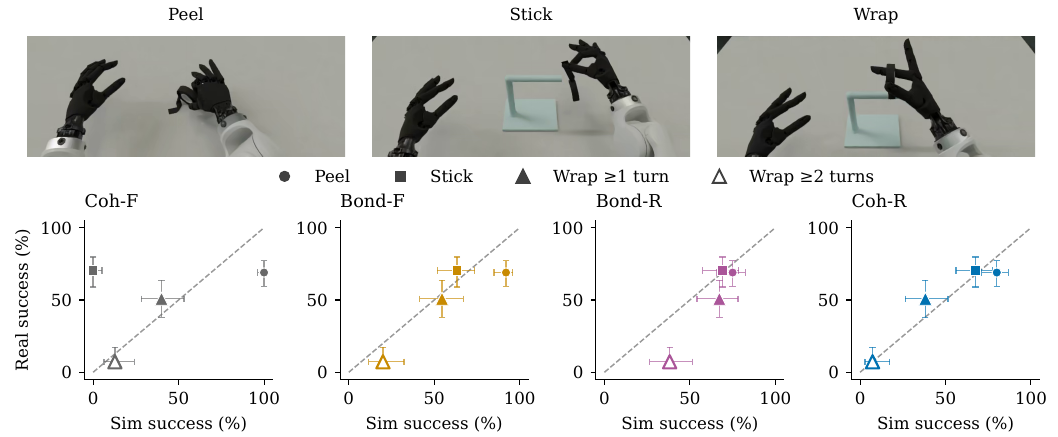}
  \caption{\textbf{Real versus simulated policy success across three tape tasks.}
Top: Coh-R rollout scenes.
Bottom: simulated success rate (horizontal) versus real success rate
(vertical); the dashed diagonal indicates equal rates.
Circles/squares denote Peel/Stick, and filled/open triangles denote Wrap
success with at least one/two completed turns.
Peel, Stick, and Wrap use 100, 71, and 55 matched cases, respectively.}
  \label{fig:policy}
\end{figure*}

Closed-loop tests ask whether the simulator predicts the same successes
and failures as the real robot. In Peel, the robot peels tape initially
adhered to the tabletop. In Stick, it picks up the tape by its free end
and attaches it to the stand. In Wrap, it inserts a finger through the
core of a roll hanging from the stand and moves the roll around the
stand, dispensing and attaching tape as it goes (\figref{fig:policy}).
The stand provides a controlled target for the Stick/Wrap wire-harnessing
subtasks.
Each policy acts on its own rollout observations, allowing tape response
to affect subsequent robot actions.

\begin{figure*}[t]
  \centering
  \includegraphics[width=0.83\textwidth]{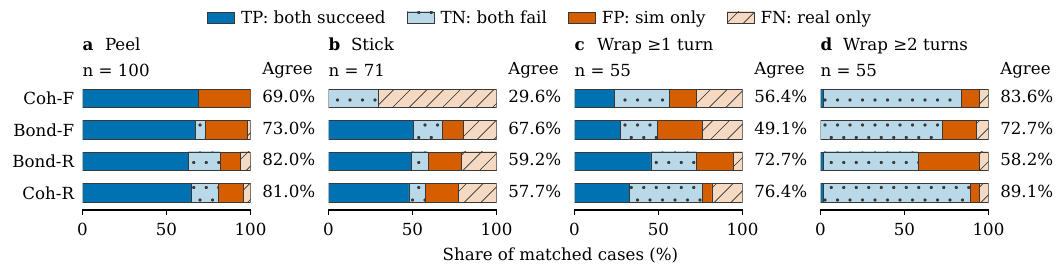}
  \caption{\textbf{Paired prediction of individual policy outcomes.}
Each bar partitions the same $n$ matched cases into four outcomes.
Blue segments indicate agreement: both succeed (TP) or both fail (TN);
orange segments indicate disagreement: sim only (FP) or real only (FN).
Right-hand values report paired agreement, $(\mathrm{TP}+\mathrm{TN})/n$.
Both Wrap thresholds use the same 55 cases.}
  \label{fig:paired}
\end{figure*}

\begin{figure}[!t]
  \centering
  \includegraphics[width=0.87\columnwidth]{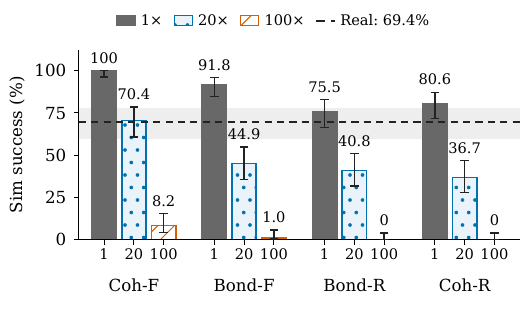}
  \caption{\textbf{Tabletop adhesion strongly affects Peel success.}
Within each model, bars show $1\times$ (solid gray), $20\times$
(blue dotted hatch), and $100\times$ (orange diagonal hatch)
tabletop $C_n$, evaluated on the same 98 cases.
Error bars show 95\% Wilson intervals; the dashed line and gray band mark real success (68/98) and its 95\% Wilson interval.
}
  \label{fig:sensitivity}
\end{figure}

\begin{figure*}[!t]
  \centering
  \includegraphics[width=\textwidth]{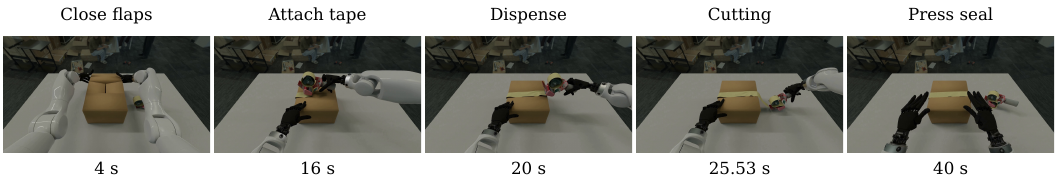}
  \vspace{-0.7cm}
  \caption{\textbf{Continuous teleoperated box sealing.}
Frames from one simulated rollout show flap closing, tape attachment,
dispensing, cutting, and pressing the seal. Times are measured from the
start of the recording.}
  \label{fig:showcase}
\end{figure*}

\subsubsection{Policies, cases, and metrics}
Training and test trials use the same tape product and similar placement regions. Initial tape position, orientation, wound-turn count, and free-end length are randomized across trials. For each real test trial, we align the simulated initial placement to the first real image using an overlay. The resulting case is shared by all four simulation configurations, pairing their starting conditions while allowing subsequent robot and tape motions to differ through policy feedback.

We fine-tune an internal vision–language–action base model on human teleoperation demonstrations to obtain one policy per task. Peel/Stick/Wrap use 284/241/241 training and 16/14/14 validation episodes. For each task, we select the checkpoint by validation action \(R^2\) and freeze it for evaluation. Real and simulated trials use the same frozen policy
weights and preprocessing.
The policy receives ego and two wrist RGB views,
proprioception, and task text, and predicts a 32-step action
chunk of end-effector poses and hand-joint commands.
Both domains use 100-Hz robot control, 30-Hz action
waypoints, and 3-Hz policy replanning.
Camera acquisition runs at 60 Hz on the real robot and
30 Hz in simulation.
Peel succeeds on complete tabletop release; Stick on lifting the tape
from the tabletop and maintaining attachment to the stand; Wrap on at
least one or two completed turns.
Peel/Stick outcomes are automatically scored and manually
verified; Wrap is manually labeled. Numerical failures count as failures.
We compare \emph{success rates} in \figref{fig:policy} and
\emph{paired agreement} in \figref{fig:paired}: the former measures
overall task difficulty, the latter whether the same cases succeed or fail.
Success-rate error bars and bands denote 95\% Wilson intervals.
To separate prediction from class imbalance, we also report balanced
accuracy (BA): the mean recall over real successes and real failures.
A constant prediction has 50\% BA.

\subsubsection{Prediction accuracy}
For Peel, both clustered variants improve individual predictions over Coh-F: agreement rises from 69\% to 81\% for Coh-R and 82\% for Bond-R, while BA rises from 50\% to 72.9\% and 76.3\%, respectively (\figref{fig:paired}). This supports clustering with either interface model.

For Stick, Coh-R and Bond-R predict similar overall success rates (67.6\% and 69.0\%, versus the real 70.4\%), but their BA values of 50.7\% and 51.7\% show weak discrimination of individual outcomes. Bond-F performs better on this measure, with 48/71 correct predictions and 64.6\% BA. Agreement in aggregate success therefore does not establish accurate case-by-case prediction.

For one-turn Wrap, Coh-R achieves 76.4\% agreement and 76.6\% BA, compared with Bond-R's 72.7\% and 72.4\%. The difference is larger at the two-turn threshold: Coh-R produces 3 false positives, versus 20 for Bond-R, while both identify only one of the four real successes. Coh-R's 49/55 agreement includes 48 joint failures; always predicting failure would agree on 51/55 cases. Thus interface rankings depend on the task, and rare two-turn
successes remain difficult to identify. 

\subsection{Sensitivity to Tabletop Adhesion}
\label{sec:exp:sensitivity}

We test how substrate-release resistance affects policy predictions by changing tabletop $C_n$ from its calibrated 0.675 to 13.5 and 67.5 ($20\times$ and $100\times$). Geometry, policy, solver settings, and hand/roll-interior adhesion remain fixed across the same 98 Peel cases.

Stronger adhesion drives the clustered models away from real success (\figref{fig:sensitivity}). At $20\times$, Bond-R's sim--real success gap grows from 6.1 to 28.6 percentage points, and Coh-R's from 11.2 to 32.7 points; at $100\times$, both fail every case. Coh-F instead reaches 70.4\% at $20\times$, close to the real 69.4\%.
This coincidence shows why matching aggregate success alone cannot calibrate adhesion. The independent peel-force measurement fixes release resistance before testing; the perturbation then exposes its effect on policy outcomes.

\subsection{Teleoperated Box Packaging}
\label{sec:exp:showcase}
\label{sec:exp:cutting}

The box-sealing demonstration uses a larger packaging-tape
asset modeled from physical measurements and scans of
Scotch 3850 tape and its dispenser (\figref{fig:showcase}).
The operator closes the flaps, anchors the free strip,
dispenses tape across the box, cuts it, and presses the seal.
The cut strip stays attached to the box while the remaining
tape stays with the dispenser.
Using Coh-R with a $1/30$-s physics time step on an NVIDIA
GeForce RTX 5090, a 3,846-step rollout advances 128.2~s of
simulated time in 304.1~s of wall-clock time, yielding
$0.422\times$ real-time throughput.

\section{Conclusion}
\label{sec:conclusion}

We presented \method, which combines a rigid wound cluster
with an advancing deformable collar for progressive tape payout.
At 32 turns, clustering gives $3.2$--$3.4\times$ mean
physics-step speedups at a fixed Newton tolerance and
$4.5$--$8.4\times$ for comparable roll motion.
Both clustered variants improve Stick replay error and Peel
outcome agreement over full-shell cohesive contact.
Bond-R provides an additional speed option, with similar
replay and Peel/Stick prediction results to Coh-R;
Coh-R performs better on Wrap.
We recommend clustering as the default representation,
with interface choice guided by task and cost.
Teleoperated box sealing demonstrates the complete workflow.


\section*{Acknowledgment}
ChatGPT assisted with drafting the manuscript.
The authors remain responsible for
all content.
\ifanonymous\else
\fi

\bibliographystyle{IEEEtran}
\bibliography{IEEEabrv,references}


\end{document}